# Building a Truly Distributed Constraint Solver with JADE

Ibrahim Adeyanju
School of Computing, The Robert Gordon University,
Aberdeen, UK

## ABSTRACT
Real life problems such as scheduling meeting between people at different locations can be modelled as distributed Constraint Satisfaction Problems (CSPs). Suitable and satisfactory solutions can then be found using constraint satisfaction algorithms which can be exhaustive (backtracking) or otherwise (local search). However, most research in this area tested their algorithms by simulation on a single PC with a single program entry point. The main contribution of our work is the design and implementation of a truly distributed constraint solver based on a local search algorithm using Java Agent DEvelopment framework (JADE) to enable communication between agents on different machines. Particularly, we discuss design and implementation issues related to truly distributed constraint solver which might not be critical when simulated on a single machine. Evaluation results indicate that our truly distributed constraint solver works well within the observed limitations when tested with various distributed CSPs. Our application can also incorporate any constraint solving algorithm with little modifications.

## 1. INTRODUCTION
Human beings in their daily activities have to make individual or collective decisions which are restricted by one or more conditions. Such real life activities can be modelled as Constraint Satisfaction problems (CSPs) and algorithms developed to give suitable solutions. A CSP comprises of a finite set of decision variables, each with a set of alternatives it can adopt and a set of constraints [1]. CSPs are solved when all the constraints between decision variables are satisfied by choices made from their domain. A distributed CSP is one in which variables and constraints are distributed among multiple agents in collaboration [2]. In such a scenario, group objectives are clearly defined but individual objectives introduce additional complexity on negotiating solutions.

This paper discusses the design and implementation of a truly distributed constraint solver using a local search algorithm on several machines. Real life applications of distributed constraint solvers include dynamic distributed resource allocation [3] which arises in problems such as distributed sensor networks, disaster rescue and hospital scheduling. Another application is building schedulers such as Distributed Meeting Scheduler and Railway Traffic regulation[4].

The rest of this paper is as follows. Section 2 provides a critical appraisal of related work while Section 3 discusses the local search algorithm used in our work. We present details of the Java Agent DEvelopment framework (JADE) in Section 4 followed by the design and implementation of our application in Section 5. Evaluation of our application is discussed in Section 6 while Section 7 concludes our work with a summary and plans for future work.

## 2. RELATED WORK
Two broad categories of techniques used in solving Constraint Satisfaction Problems (CSP), centralised or distributed, are Exhaustive Search and Local Search. Exhaustive Search also known as Systematic Backtracking involves starting with a partial solution that is carefully chosen and incrementally searching through all the possible combination of different values of the variables until a complete solution that satisfies all constraints is found. Exhaustive search algorithms are guaranteed to find one or more solutions if they exist or could determine if no solution exists at all. Current backtracking algorithms include Back-jumping schemes [5,6], Asynchronous Weak Commitment search [7] and Asynchronous Forward Checking [8].

Local search involves starting with a partial solution through random assignment of values to variables involved in a CSP. An improvement in the random solution is then sought through successive iterations by exploring different points in the search space until a valid solution is found or the maximum time allowed has elapsed. Simulated Annealing [9,10], Breakout Algorithm [11], Tabu Search [12], Distributed Breakout Algorithm [13,14] and Distributed Penalty-driven Local search algorithm [15] are examples of existing local search algorithm for constraint solving.

Distribution of CSPs across multiple machines, rather than simulation on a single machine, requires the development of a distributed system. Several technologies exist for building such distributed systems. These include Remote Procedure Calls (RPC) [16,17], .NET Remoting [18,19], Remote Method Invocation (RMI) [20,21], Common Object Request Broker Architecture (CORBA) [22,23] and Simple Agent Communication Infrastructure (SACI) [24,25] among others.

In our work, we used Java Agent DEvelopment Framework (JADE) [26,27], an open source platform for peer-to-peer agent applications. We chose JADE because it is open-source and used more widely to build multi-agent systems.

## 3. CONSTRAINT SOLVER
Our distributed constraint solver application is built using a local search algorithm called Distributed Penalty-driven Local search (DisPeL) [15] as its underlying constraint solver. DisPeL is a local search algorithm for solving distributed CSPs, where each agent controls just one variable, by finding the first solution that satisfies all constraints simultaneously. Collaborating agents take turns in a fixed ordering to improve a random initialization. Gradual sequential improvements are found iteratively rather than the best possible improvement as in conventional hill-climbing algorithms. This causes a reduction in communication costs since all improvements are accepted and the information used in making decisions is always coherent [15].

DisPeL's core strategy is in its use of two types of penalties (temporary and incremental) in resolving deadlocks (local optima) by modifying the underlying cost landscape. Deadlocks occur when the solution to a CSP cannot be improved further by agents although no suitable solution has been obtained. Penalties are used to locally perturb the solution thereby forcing the agents to try other combination of values through exploration of other areas in the search space. Temporary penalties are removed immediately after use while incremental penalties are only reset when current value of an agent does not violate any of its constraints. We use the stochastic version of DisPeL where the decision on whether to use temporary or incremental penalties is done randomly.

Penalties are used collaboratively. When an agent detects a deadlock and has to use a penalty, it implements the penalty on its current value and requests all neighbours with a lower ordering priority (for incremental penalty) or those neighbours with a lower ordering priority and violate constraints with self (for temporary penalty) to implement the same penalty on their current value assignments. It is also assumed that all constraints are uni-directed; therefore each agent in DisPeL will locally evaluate all constraints attached to its variable. Hence, each agent will communicate in a synchronised manner with all other agents that are co-constrained with it exchanging value assignments and requests to impose penalties [15].

## 4. JADE FRAMEWORK

Distributed constraint solvers are multi-agent systems since they attempt to find suitable solutions to CSPs whose agents are distributed across several locations. Agent-oriented applications combine artificial intelligence with distributed system techniques by modelling components as agents. Each agent is autonomous, proactive, and has the ability to communicate with other agents to achieve personal and communal goals [26]. Such applications have a peer to peer architectural model where any agent is able to send or receive communication from any other agent within the application. Open source middleware which provide domain-independent infrastructure can facilitate communication in multi-agent systems thereby allowing application developers to focus on production of the business logic.

Java Agent DEvelopment framework (JADE) is a completely distributed middleware system with a flexible infrastructure that allows easy extension [6]. The framework facilitates development of complete agent-oriented applications by means of a run-time environment implementing the life-cycle support features required by agents and the core logic of agents themselves among other tools. JADE is a software platform written in Java that provides basic middleware-layer functionalities which are independent of the specific application and which simplify the realization of distributed applications that exploit the software agent abstraction [26]. Each agent in JADE complies with the FIPA (Foundation for Intelligent Physical Agents) specifications and therefore has such basic qualities as autonomy, pro-activeness, responsiveness, and social ability with secondary qualities like mobility, adaptability and rationality. Any multi-agent system based on JADE is loosely coupled, peer-to-peer and message communication between agents are asynchronous. Each agent has its own thread of execution using this to control its life cycle and decide autonomously to perform specific tasks.

### 4.1 JADE Architecture

A JADE platform consists of a runtime environment (also called containers) that can be distributed over the network and provides all the services needed for hosting and executing agents. A special container, called the *main container* must always be active in a platform and all other normal containers register with it as soon as they start and must therefore know the main container's host address and port. A diagram showing the typical architecture of the JADE platform is shown in Figure 1. Starting another main container elsewhere in the network constitutes a different platform to which new normal containers can possibly register. The main container manages the container table (CT), which is the registry of object references and transport addresses of all container nodes in the platform; manages the global agent descriptor table GADT), which is the registry of all agents present in the platform, including their current status and location; and hosts the AMS (Agent Management Service) and DF (Directory Facilitator), the two special agents that provide the agent management service and the default yellow page service of the platform respectively. The DF is not used in our work since the number of agents involved in solving the distributed constraint problem does not change throughout the solution finding process.

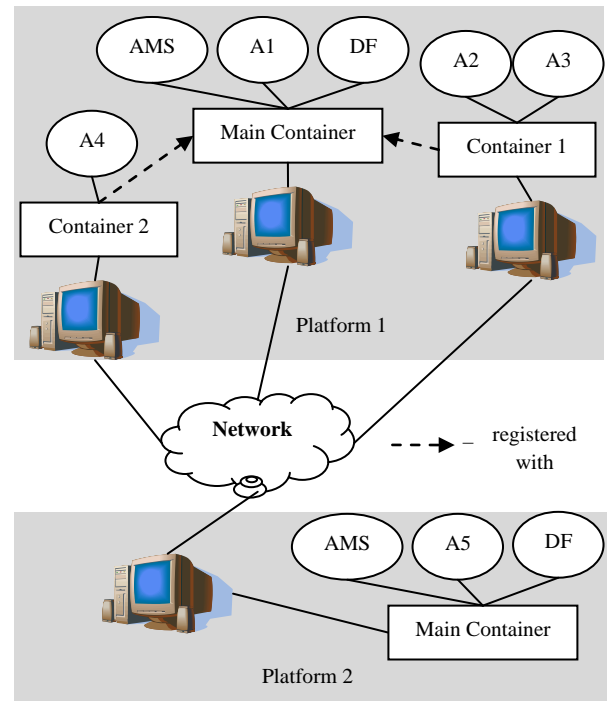

**Figure 1. Relationship between Containers and platforms on the JADE architecture [27]**

Because agent communication is peer-to-peer, each agent maintains a local agent descriptor table (LADT) which it searches first when communicating with any other agent and only involves the main container's GADT if the agent's address is not on its LADT and caches it locally for future use. Agents in JADE are identified by a globally unique name called an Agent Identifier (AID) consisting basically of the agent's local name and its addresses (usually inherited from the platform) [26]. Each agent can communicate transparently regardless of their actual location: same container (e.g. A2 & A3 in Figure 1), different containers in the same platform (A1 & A2) or different platforms (A4 & A5) provided they know each other's agent identifier [6].

## 4.2 Message Transport Service

JADE includes a Message Transport Service (MTS) that manages all message exchange within and between platforms. All standard Message Transport Protocols (MTPs) defined by FIPA are implemented by this service to promote interoperability between different non-JADE platforms. Each MTP includes the definition of a transport protocol and a standard encoding of the message envelope. HTTP-based MTP are always started by default with the initialization of a main container while no MTP is activated on normal containers. This creates a server socket on the main container host and listens for incoming connections over HTTP at the URL specified. Whenever an incoming connection is established and a valid message is received over that connection, the MTP routes the message to its final destination which, in general, is one of the agents located within the distributed platform [26]. The platform uses a proprietary transport protocol called IMTP (Internal Message Transport Protocol) internally to perform message routing for both incoming and outgoing messages using a single-hop routing table that requires direct visibility among containers. IMTP is also used to transport internal commands needed to manage the distributed platform as well as monitor the status of remote containers. The two main implementations of IMTP available are Java RMI which is the default option and a proprietary protocol using TCP sockets that circumvents the absence of Java RMI in the J2ME environments. The default Java RMI implementation was used throughout the development of our distributed constraint solver.

## 4.3 Agent Tasks - Behaviour scheduling

An agent in JADE carries out its tasks within program elements called "behaviours". A behaviour represents a task that an agent can carry out. An agent can execute several behaviours concurrently although the scheduling of behaviours in an agent is not pre-emptive but cooperative. This means that when a behaviour is scheduled for execution its *action()* method is called and runs until it returns. Therefore it is the programmer who defines when an agent switches from the execution of a behaviour to the execution of the next one. When there are no behaviours available for execution the agent's thread goes to sleep in order not to consume CPU time and is woken up as soon as there is a behaviour again available for execution [27].

## 4.4 Agent Communication

The communication paradigm adopted in JADE is the asynchronous message passing [26]. Each agent has a message queue where the JADE runtime posts messages sent by other agents; whenever a message is posted in the message queue, the receiving agent is notified. The programmer however determines if and when the agent actually picks up the message from the queue to process it. This process is shown in Figure 2. The format of messages in JADE is compliant with FIPA-ACL message structure specifications and has fields such as the sender, list of receivers, communicative act (REQUEST, INFORM, PROPOSE etc), content, content language and ontology.

## 5. DESIGN & IMPLEMENTATION

The design of our truly distributed constraint solver application is discussed under two sections via the *User Interface* (Section 5.1) and the underlying *Distributed Constraint Solver* (Section 5.2). The user interface is that part of the application that interacts with the user by allowing the user to input values and displays the final result of the computation. The underlying distributed constraint solver deals with the structure of the algorithm and how it was integrated into the JADE platform that was used to allow for true distribution of the agents on different machines in solving constraint satisfaction problems.

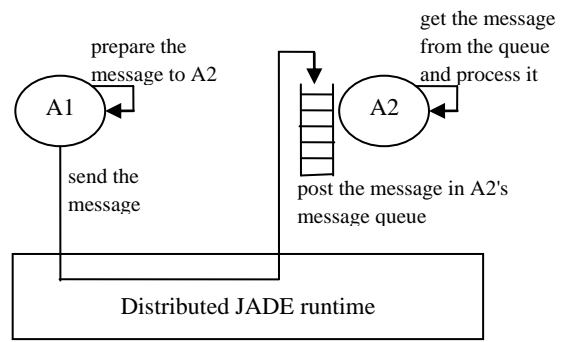

**Figure 5. JADE asynchronous message passing paradigm**

### 5.1 Graphical User Interface (GUI)

The graphical user interface would allow a user to input some values for an agent involved in the DisCSP before it is active and starts communicating with other agents in order to solve the problem. The results are shown on a dialog box. The graphical user interface for the distributed constraint solver is shown in Figure 3. This user interface can be used only once by a *single* agent involved in solving a DisCSP in collaboration with other agents; the application *must* be restarted in order to solve any other DisCSP. The application can only be exited using the "Exit Application" button or the "Exit Application" menu item under the "File" menu which shows a confirmation dialog before exiting. An error dialog box is popped up if any of the input fields contain an invalid value and the information on the dialog gives a hint as to the possible cause of the error.

The "Start Agent" button is used to activate an agent involved in solving a DisCSP in collaboration with other active agents on the same main container though they might be on different machines. The agent would only be started if all input fields are found to be correct from all validation checks done. This button is disabled afterwards to ensure that another agent cannot be started using this application instance. The "Clear All Inputs" button is used to clear all inputs entered before the agent is started and is also disabled once the agent is started as no input adjustments are allowed afterwards.

The "Start Main Container on this Host" button is used to start the main container on the same host as the agent and must be used before any normal agent is started. The menu item is disabled if the main container is started successfully or an agent is started on the same host. Care must be taken to click this button on just one of the machines that are involved in solving a DisCSP because there would be no communication between the agents if each agent host starts its own main container. The "Exit Application" button exits the application displaying a confirmation dialog box and the program only exits if no active agent is using the interface; in other words, no agent is started yet or a DisCSP has been solved (partially or fully) and agent's communication with others is complete.

The "Help" menu contains a single menu item ("About") which pops up information about the author, the version and copyright notices. Basic instructions to guide the user to enter correct and valid inputs values are shown on the upper section of the interface. This ensures that the user reads them first before entering values in the input boxes provided in the lower section of the user interface. The *Agent's variable name* is assumed to be same as the agent's name since each agent has only one variable. Validation checks are also done to ensure that this input field is not blank before the agent is started. The *Agent's domain value* field takes the domain values for this agent's variable which must be separated by comma. We restricted valid domain value type for this application to Integers. The *Maximum Iterations* input field takes an Integer value that indicates the cycle after which the agent would stop communicating with neighbouring agents in the DisCSP if a solution has not been found.

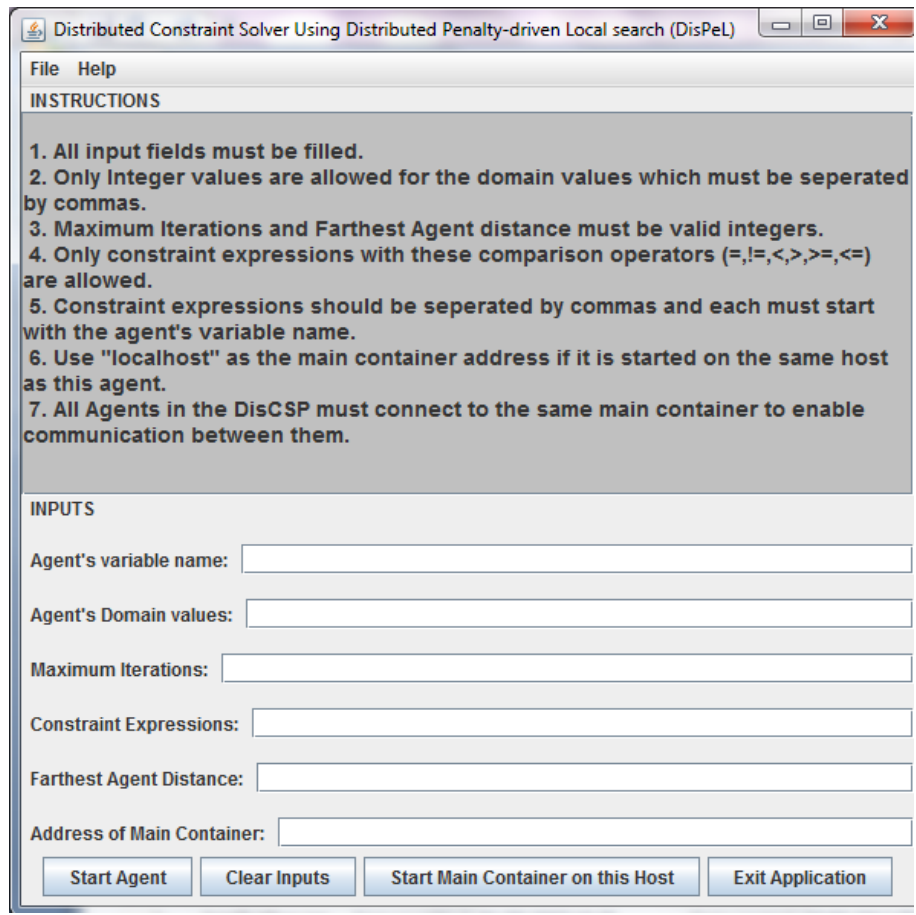

**Figure 3. User Interface for the Distributed Constraint Solver**

The *Constraint Expressions* field allows the user to input the constraints between the agent and other agents in form of mathematical expressions. Our application accepts only comparison (>, <, !=, =, >=, <=) constraints expression to between just *two variables* i.e. the agent and any another agent. Each constraint expression starts with the agent's variable name followed by the operator, then the other agent's variable name. Each constraint expression must as well be separated by a comma if the agent has constraints with more than one agent.

The *Farthest Agent Distance* field accepts an Integer value that is used for termination detection by ensuring that all agents in the DisCSP have obtained solutions to their local problems. The value gives an estimate of the number of agents in between the two farthest agents in the DisCSP that do not have direct constraints together but are indirectly connected through other agents. The *Address of Main Container* field takes a string value that indicates the address (usually HTTP address, fully qualified with computer name and domain name) where the main container that routes messages between JADE agents is located. The main container must be started first before other agents can join the JADE platform when trying to solve a DisCSP. The loop-back address of the host ('localhost') could be used if the agent is starting on the same machine as the main container.

## 5.2 Constraint solving technique

The Distributed Constraint Solver (DCS) underlying our GUI discussed in Section 5.1 is based on the Distributed Penalty-driven Local search (DisPeL) [15] algorithm. Here, we discuss the implementation of a DisPeL DCS with JADE to ensure true distribution of the agents on different machines. It should be emphasized that all versions of DisPeL were previously simulated on single machines by the original author and our implementation of a true DCS is novel. Real distribution of DisCSPs leads to other important research problems. The problems we encountered while implementing a DisPeL DCS on several machines are in next Sub-Sections.

### 5.2.1 Ownership of the DisCSP

When DisPeL algorithm was simulated to solve DisCSPs on a single machine; all the constraint expressions were either randomly generated [15] or entered through the same GUI [29]. It was relatively easy to identify constraint expressions for each variable. However, to run DisPeL on several machines, we have to address the problem of who initialises the DisCSP and sorts out all the constraint expressions. Ideally, each agent involved in the solving a DisCSP should only know about other agents that it has constraints with but the algorithm solves the problem based on the fact that the whole DisCSP is known from the beginning before starting the process towards finding a suitable solution and that the DisCSP is static (unchanged) throughout until a solution is found or the maximum iterations allowed is reached. Some important features of the DisPeL algorithm such as ordering all variable names lexicographically or using the Distributed Agent Ordering scheme is based on this fact and might be difficult to circumvent. We address this problem by ensuring that each agent only knows about the agents it is constrained with and its local CSP are entered manually from the agent's GUI. Each agent also has an estimate of the size of the whole DisCSP from farthest agent distance entered on its GUI. We also assumed that all agents involved in the DisCSP are started near simultaneously to ensure that the problem is complete before the process towards finding a solution starts.

*5.2.2 Constraints and global constants validation*
Also closely related to the problem of DisCSP ownership is the validation of constraint expressions and global constants like the *maximum number of iterations*. This does not pose a problem when the constraint solver is implemented on a single machine with single interface for taking all inputs. For instance, if agent A and B have constraints "A>B" between them, it must be ensured that the same equivalent expression ("A>B" on agent A's interface and "B<A" on agent B) is typed on their interfaces to avoid conflicts. Global constants like the "maximum number of iterations" should also be identical for all agents as used in the original algorithm. To address this problem, we validate equivalent expressions across constrained agents by passing the constraint operators with other messages that were communicated between agents. Conflicting constraints are then ignored in the process of finding a solution to the DisCSP. We did not think having the same values of the "maximum number of iterations" was critical to finding a solution to our kind of DisCSPs since this could mean stopping all agents if any of them has a different value from any other one. Therefore, we did not validate the "maximum number of iterations" input across agents.

*5.2.3 Termination Detection*
The constraint solving process has to be terminated when all of the agents obtain solutions to their constraints. Such termination detection is relatively easy when all agents are implemented on a single machine and have a single application entry since each agent can be checked to have obtained a solution before the application is terminated. The termination detection is more complex when agents are situated on different machines with multiple application entries. We address this problem as suggested by original author of our constraint algorithm [15] by using the same method as Distributed Breakout Algorithm [13].

*5.2.4 Unreliable network communication*
Since the agents involved in the DisCSP could be located on different machines connected through a computer network (local network preferably), the issues of communication delays, network congestion, packet corruption and time-outs are also paramount. There would be no cause to consider this issue when all the agents are on the same machine. We used the remedy suggested by the author of our constraint solver [15] where agents are allowed to resume activity if messages have not been received after a reasonable amount of time. Agents in such situation assume that their neighbours' values are unchanged.

## 6. EVALUATION
We tested extensively our truly distributed constraint solver application with sample DisCSPs. Test cases were designed based on the functional requirements of our application. Seven DisCSPs having up to a maximum of four agents were formulated for this purpose. The formulated DisCSPs are shown in Table 1. The diameter of agent network was taken as the farthest agent distance parameter used in termination detection. This can be obtained by drawing the agent tree for a DisCSP and counting the number of agents from the top to the bottom of agent tree as illustrated in Figure 4.

We observe from Table 1 that all the tested DisCSPs gave correct outputs with the only one not solved showing the interim results when the maximum iteration was reached. There is sometimes a variation in the number of iterations reached across constrained agents when a final solution was obtained because of the asynchronous nature of the JADE. A screenshot from one of the agent's GUI during our evaluation for test case 7 is shown in Figures 5.

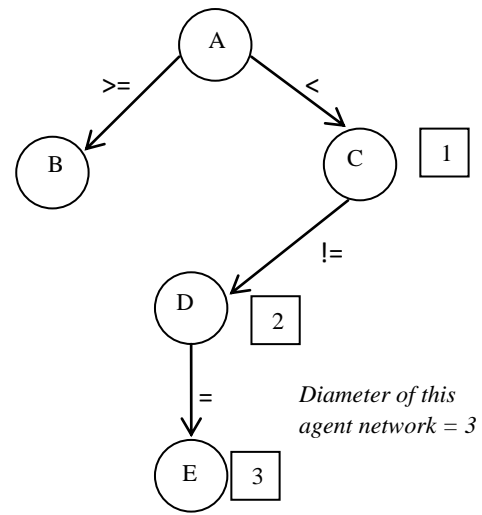

**Figure 4. Calculating agent network diameter**

## 7. CONCLUSION
This paper discussed our work on the development of a truly Distributed Constraint Solver application based on a local search algorithm (DisPeL) on different machines. The JADE framework class libraries were used to implement a multi-agent system that enables the true distribution of CSPs.

We intend to extend our application to allow each agent have multiple variables in addition to being truly distributed on several machines. The type of constraint expressions handled by our software will also be extended to allow Comparison, Boolean and Arithmetic operations between more than two variables. Variables other than Integers like Double, String and other objects like Date that would be more useful in real life applications will also be considered in future version of our application. Finally, we would consider distributing the agents in our constraint solver over a wide area network other than a local network.

**Table 1 DisCSPs formulated for testing our application and results**

| Test case | Agents and their domain values | Constraint expressions | Maximum Iterations | Network diameter | Final Results | Iterations used | Remarks |
|---|---|---|---|---|---|---|---|
| 1 | A {1,2,3,4,5} <br> B {2,4,6,8,10} | A>B | 100 | 1 | A= 4 <br> B= 2 | 3 | Final Solution |
| 2 | A {1,2,3,4,5} <br> B {6,7,8,9,10} | A=B | 100 | 1 | A= 2 <br> B= 6 | 100 | Interim Solution |
| 3 | A {1,2,3,4,5} <br> B {2,4,6,8,10} <br> C {1,3,5,7,9} | A>B <br> A<C | 100 | 2 | A= 5 <br> B= 4 <br> C= 9 | 6 <br> 6 <br> 4 | Final Solution |
| 4 | A {1,2,3,4,5} <br> B {2,4,6,8,10} <br> C {1,3,5,7,9} | A!=B <br> A<C <br> B>C | 100 | 2 | A= 1 <br> B= 6 <br> C= 3 | 6 | Final Solution |
| 5 | A {1,2,3,4,5} <br> B {2,4,6,8,10} <br> C {1,3,5,7,9} <br> D {6,7,8,9,10} | A=B <br> A!=C <br> B!=D <br> C>D | 100 | 2 | A= 2 <br> B= 2 <br> C= 7 <br> D= 6 | 6 | Final Solution |
| 6 | A {1,2,3,4,5} <br> B {2,4,6,8,10} <br> C {1,3,5,7,9} <br> D {6,7,8,9,10} | A=B <br> B>C <br> C<=D | 100 | 3 | A= 2 <br> B= 2 <br> C= 1 <br> D= 8 | 7 <br> 8 <br> 9 <br> 8 | Final Solution |
| 7 | A {1,2,3,4,5} <br> B {2,4,6,8,10} <br> C {1,3,5,7,9} <br> D {6,7,8,9,10} | A<B <br> A>C <br> A<=D <br> B=D <br> C!=D | 100 | 3 | A= 4 <br> B= 6 <br> C= 3 <br> D= 6 | 8 | Final Solution |

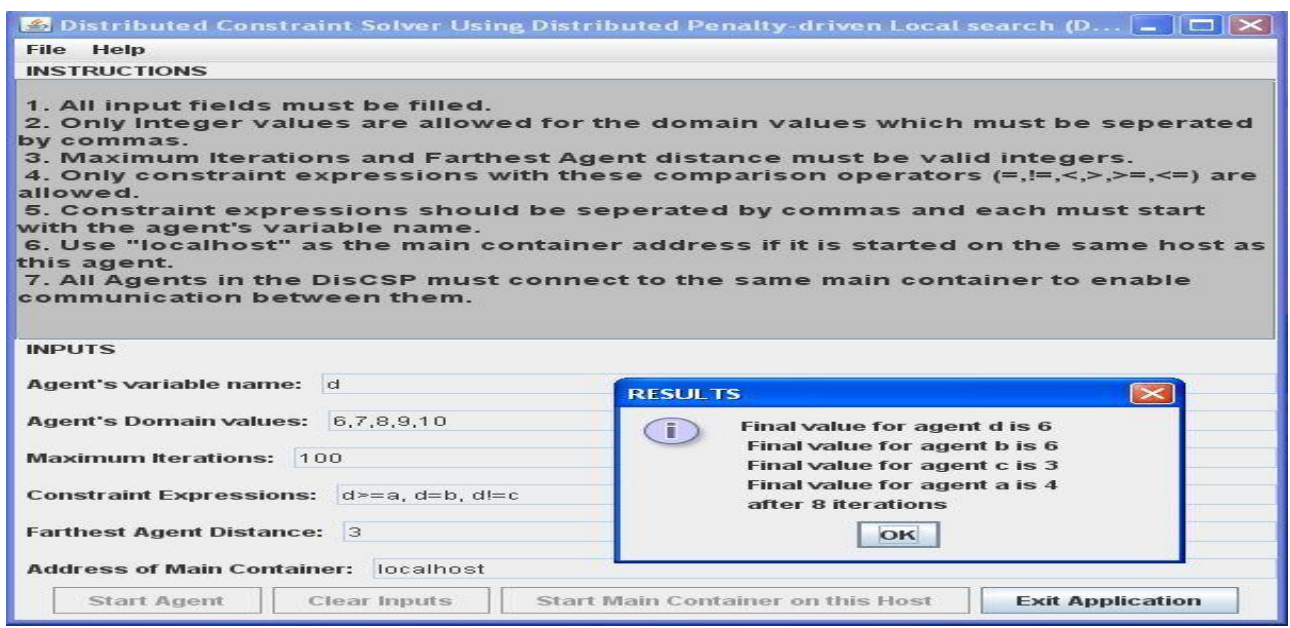

**Figure 5. Agent D's GUI screenshot in Test case 7**

## 8. ACKNOWLEDGEMENTS
The author is grateful to Dr. Hatem Ahriz of The Robert Gordon University, Aberdeen for his very useful feedback.